\documentclass[runningheads,a4paper]{llncs}

\usepackage{amssymb}
\usepackage{amsmath} 
\usepackage{graphicx}
\usepackage{epstopdf}
\usepackage{multirow}
\usepackage{gensymb}
\usepackage{cite}

\usepackage{url}
\urldef{\mailsa}\path|{p.proenca, yang.gao}@surrey.ac.uk|    
\newcommand{\keywords}[1]{\par\addvspace\baselineskip
\noindent\keywordname\enspace\ignorespaces#1}

\begin{document}

\mainmatter  

\title{Probabilistic Combination of Noisy Points and Planes for RGB-D Odometry}

\titlerunning{Probabilistic Combination of Noisy Points and Planes for RGB-D Odometry}

%
%
\author{Pedro F. Proen\c{c}a%
\and Yang Gao}

\institute{Surrey Space Centre, Faculty of Engineering and Physical Sciences,\\ 
University of Surrey, Guildford, UK\\
\mailsa\\}

%
%

\toctitle{Probabilistic Combination of Noisy Points and Planes for RGB-D Odometry}
\tocauthor{Authors' Instructions}
\maketitle

\begin{abstract}
This work proposes a visual odometry method that combines points and plane primitives, extracted from a noisy depth camera. Depth measurement uncertainty is modelled and propagated through the extraction of geometric primitives to the frame-to-frame motion estimation, where pose is optimized by weighting the residuals of 3D point and planes matches, according to their uncertainties. Results on an RGB-D dataset show that the combination of points and planes, through the proposed method, is able to perform well in poorly textured environments, where point-based odometry is bound to fail.
\keywords{Visual odometry $\cdot$ Depth cameras $\cdot$ Uncertainty propagation $\cdot$ Probabilistic plane fitting}
\end{abstract}

\section{Introduction}

Historically, the problem of visual odometry and SLAM \cite{PTAM} has been mostly addressed by using image feature points. However, low-textured environments pose a problem to such approaches, as interest points can be insufficient to estimate precisely the camera motion and map the environment. Thus, other primitives (e.g. planes and lines) become more relevant, particularly in indoor environments, where planar surfaces are predominant. \par
The geometry of indoor environments can be unambiguously captured by active depth cameras, which are capable of capturing dense depth maps, at 30 fps, regardless of the image textures. Therefore, the combination of such depth maps with RGB images, known as RGB-D data, has led to the emergence of several robust visual odometry methods. Likewise, in this work, we propose a visual odometry method that uses both points and planes, extracted from RGB-D. While Iterative Closest Point (ICP) has been the standard approach \cite{PlaneAndsurfels_2014} to exploit dense 3D information, plane primitives are an attractive alternative, as they can be handled more efficiently during pose estimation \cite{PointsAndPlanes_2013} and be used for mapping to yield highly compact representations \cite{PlaneAndsurfels_2014}. In the context of the former, we introduce a novel plane-to-plane distance, as an alternative to the typical point-to-plane distances.
\par
Due to the systematic noise of these depth sensors, our method models the uncertainty of the 3D points and planes in order to estimate optimally the camera pose. This is done by propagating the depth ucertainty to the pose optimization, so that pose is estimated by minimizing the distances between feature matches, that are scaled according to the 3D feature geometry uncertainties. The intuition behind this probabilistic framework is that points and planes that are far from the camera should have less impact on the pose estimation than closer ones and the weights of estimated plane equations should depend on the number and distribution of point measurements on the planes. \par 
Results on a public RGB-D dataset show the benefit of modelling uncertainty, for a structured-light camera, and show more robustness, in low-textured scenes, by combining points and planes rather than just using points.

\section{Related Work}

Several works have recently addressed RGB-D SLAM by using plane primitives. Trevor et al. \cite{trevor2012planar} proposed a planar SLAM that used data from 2D laser scans and depth cameras as they complement each other in terms of field of view and maximum range. Renato et al. \cite{PlaneAndsurfels_2014} proposed mapping the environment using bounded planes and surfels to represent both planar and non-planar regions. Taguchi et al. \cite{PointsAndPlanes_2013} avoided the geometric degeneracy of planes by proposing the combination of 3D points and planes through both a closed-form solution and bundle adjustment. Kaess \cite{kaess2015simultaneous} proposed a minimal plane parameterization more suitable for least-squares solvers than the frequently used Hessian normal form. Ma et al. \cite{ma2016cpa} proposed combining a global plane model with direct SLAM to reduce drift. 
\par
The combination of 3D lines and points has also shown to be advantageous for RGB-D odometry, by Lu and Song \cite{PLVO}, who proposed taking into account the depth uncertainty of a structured-light camera, by modelling the uncertainty of 3D line and point extraction and optimizing simultaneously the camera pose and the 3D coordinates of the primitives through maximum likelihood estimation. Moreover, the uncertainty of plane extraction has been analyzed by Pathak et al. \cite{pathak2010uncertainty}, who compared several direct and iterative plane fitting methods, in terms of accuracy and speed.

\begin{figure}
\centering
	\includegraphics[scale=0.355]{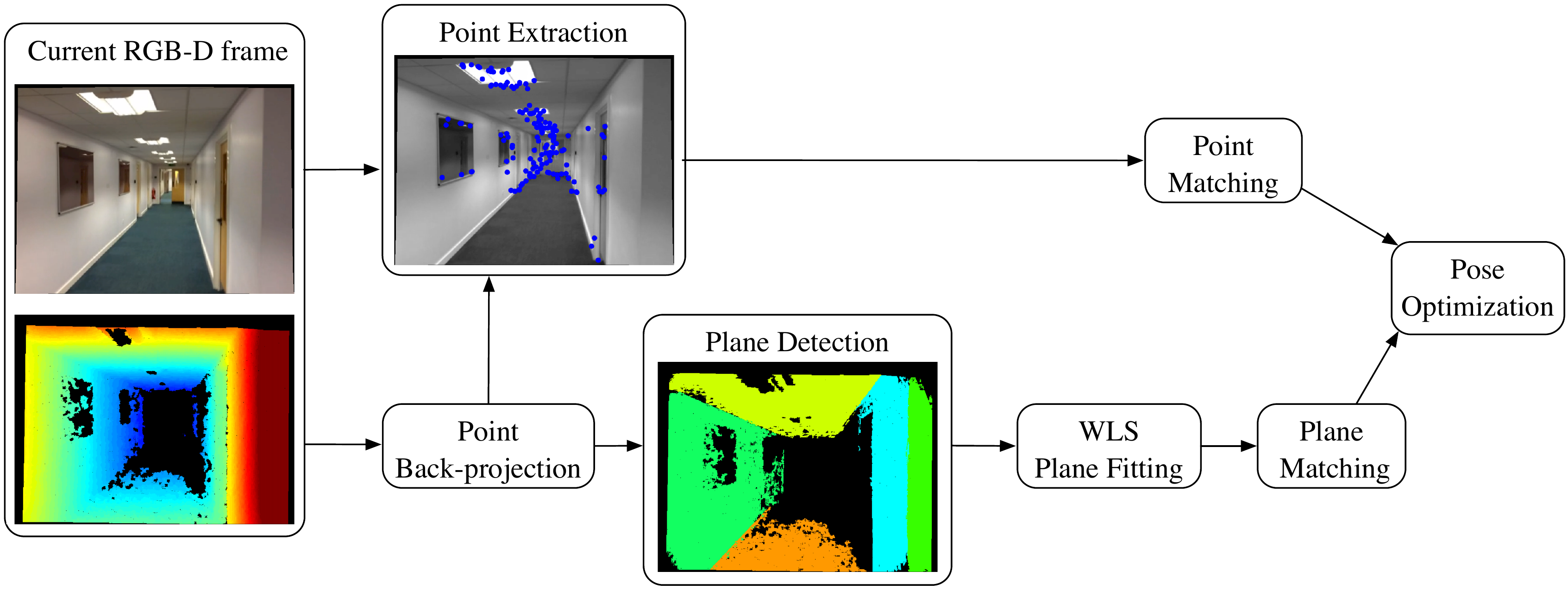}
	\caption{System overview}
	\label{fig1}
\end{figure}

\section{System Overview}
The proposed system, outlined in Fig. \ref{fig1}, starts by detecting points and planes from an RGB-D frame. Samples from the noisy depth map are used to obtain 3D points, through back-projection, and the 3D points corresponding to detected planes are used in turn to estimate the plane parameters through a weighted least squares framework, which takes into account the depth measurement uncertainties. Once, point and plane matches are found between two adjacent frames, pose is estimated in iteratively reweighted least squares by minimizing both the point and plane residuals, according to their uncertainties. For this purpose, uncertainty is propagated throughout this process. The modules of the system and the derivation of uncertainty are described in further detail below.

\section{Point Extraction and Back-Projection}

Image points are detected by relying on SURF features \cite{bay2006surf}. Given a calibrated RGB and depth image pair, the 3D coordinates: $P=\begin{bmatrix}X, Y, Z\end{bmatrix}^\top$of a detected image point $p=\{u,v\}$ can be directly obtained by back-projecting the respective depth pixel value $Z$:

\begin{equation}
\label{eq:backprojection}
P = Z\begin{bmatrix}(u - c_x)/f_x \\ (v - c_y)/f_y \\ 1 \end{bmatrix}
\end{equation}
where $\{f_x,f_y\}$ and $\{c_x,c_y\}$ are respectively the focal length and principal point of the RGB camera. As shown by \cite{PLVO}, the uncertainty of the 3D point coordinates: $\Sigma_P$ can be obtained by the first order error propagation of (\ref{eq:backprojection}):

\begin{equation}
\label{eq:backprojection_uncertainty}
\Sigma_P = J_P \begin{bmatrix}\sigma^2_p & 0 & 0\\0 & \sigma^2_p & 0 \\0& 0 & \sigma_Z^2\end{bmatrix}J_P^\top
\end{equation}
where $J_P$ is the Jacobian of (\ref{eq:backprojection}) with respect to $p$ and $Z$, $\sigma^2_p$ is the pixel uncertainty and $\sigma_Z^2$ is the uncertainty of the depth value. While it is generally accepted that a $\sigma_p =1/2$ approximates the pixel quantization error, the depth measurement uncertainty depends on the type of sensor used. Since structured-light cameras (e.g. Kinect V1) were used in our experiments, we adopt the theoretical model of \cite{Khoshelham2012}: $\sigma_Z = 1.425\times10^{-6} Z^2 \ [mm]$, which addresses the depth quantization of this type of sensors.

\section{Plane Extraction}

Planes are first detected by using the method of \cite{feng2014fast}, which processes efficiently organized point clouds in real-time, yielding a segmentation output, as the one shown in Fig. \ref{fig1}. However, the detected planes may contain outliers, thus RANSAC is used additionally to filter each detected plane. Although, plane fitting is already performed by this RANSAC process, we use the method proposed below to obtain the plane parameters and derive their uncertainty.

\subsection{Plane Fitting through Weighted Least Squares}

It is useful to express planes as infinite planes in the hessian normal form: $\theta =\{N_x,N_y,N_z,d\}$. However, such a representation is overparameterized, thus the estimation of these parameters by unconstrained linear least squares is degenerate. This issue has been solved in \cite{pathak2010uncertainty} by using constrained optimization and in \cite{weingarten2004probabilistic} by using a minimal plane parameterization. Similarly to \cite{weingarten2004probabilistic}, we use a minimal plane representation: $\theta_m = \begin{bmatrix}N_x,N_y,N_z\end{bmatrix}/d$, as an intermediate parameterization. Since, a plane with $d=0$ implies detecting a plane that passes through the camera center (i.e. projected as a line), it is safe to use this parameterization. The new parameters are then estimated by minimizing the point-to-plane distances through the following weighted least-squares problem:
\begin{equation}
\label{eq:lsq_fitting}
E = \sum_{i=1}^{N} \frac{w_i(\theta_mP_i +1)^2}{2}
\end{equation}
where the scaling weights were chosen to be the inverse of the point depth uncertainties: $w_i = \sigma_Z^{-2}$, although other choices are also found in literature \cite{pathak2010uncertainty}. By setting the derivative of (\ref{eq:lsq_fitting}), with respect to $\theta_m$, to zero, we arrive at the solution of the form: $\theta_m^\top = A^{-1}b$, where:

\begin{equation}
\label{eq:plane_fitting_solution_A}
A = \begin{bmatrix}  
\sum_{i=1}^{N}w_iX_i^2 &   \  \sum_{i=1}^{N}w_iX_iY_i  & \ \sum_{i=1}^{N}w_iX_iZ_i \\[0.5em]
\sum_{i=1}^{N}w_iX_iY_i &   \  \sum_{i=1}^{N}w_iY_i^2  &  \ \sum_{i=1}^{N}w_iY_iZ_i \\[0.5em]
\sum_{i=1}^{N}w_iX_iZ_i &   \  \sum_{i=1}^{N}w_iY_iZ_i  &  \  \sum_{i=1}^{N}w_iZ_i^2 
\end{bmatrix} 
\end{equation}
\\
\begin{equation}
\label{eq:plane_fitting_solution_b}
b = -\begin{bmatrix}  
\sum_{i=1}^{N}w_iX_i\\[0.5em]
\sum_{i=1}^{N}w_iY_i \\[0.5em]
\sum_{i=1}^{N}w_iZ_i
\end{bmatrix} 
\end{equation}
and the covariance of $\theta_m$ is given by the inverse Hessian matrix, i.e., $\Sigma_{\theta_m} = H^{-1}$ where $H$ is simply $A$. The Hessian normal form can then be recovered by:

\begin{equation}
\label{eq:plane_params}
\theta = \frac{\begin{bmatrix} \theta_m & 1\end{bmatrix}}{\|\theta_m\|}
\end{equation}
and the respective uncertainty is obtained via first order error propagation: $\Sigma_{\theta}  = J_{\theta}  \Sigma_{\theta_m} J_{\theta}^\top$, where $J_{\theta}$ is the Jacobian of (\ref{eq:plane_params}).

\section{Point and Plane Matching}
\label{matching}

Both point and plane feature matching capitalize on small frame-to-frame motion. Point correspondences are estabilished between consecutive frames by matching the feature descriptors through a $k$-NN descriptor search. Given a set of putative matches $k$ to point $p$, we select the closest match, whose point coordinates lie in a circular region defining the neighbourhood of $p$. \par
For matching planes, we first obtain 1-to-N candidate matches by enforcing the following constraints:
\begin{itemize}
\item Projection overlap: The projections of two planes, defined as the image segments covered by the inliers of the planes, must have an overlap of at least 50\% the number of plane inliers of the smallest plane. This can be checked efficiently by using bitmask operations.
\item Geometric constraint: Given the Hessian plane equations of two planes: $\{N,d\}$ and $\{N',d'\}$, the angle between the plane normals: $\arccos(N \cdot N')$ must be less than 10$^\circ$ and the distance: $\lvert d-d'\rvert$  must be less than 10 cm.
\end{itemize}

\begin{figure}
\centering
	\includegraphics[scale=0.6]{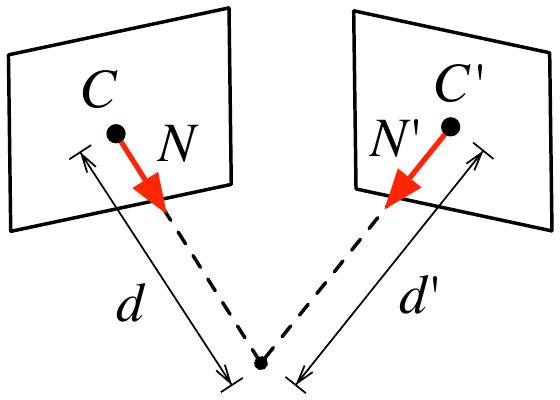}
	\caption{Geometry of two planes and their representation as points: $\{C,C'\}$}
	\label{fig2}
\end{figure}

To select the best plane match between the plane candidates, we introduce here the concept of plane-to-plane distance, so that the plane candidate with the minimum plane-to-plane distance is selected. Let $\{N',d'\}$ and $\{N,d\}$ be the equations of two planes then the distance between the two planes is expressed by:

\begin{equation}
\label{eq:plane_dist}
\lVert C - C' \rVert = \lVert d'N' - dN \rVert
\end{equation}
where $C$ and $C'$ represent points on the planes, as shown in Fig. \ref{fig2}.

\section{Pose estimation}
To estimate the rigid body transformation $\{R,t\}$ between frames, we minimize simultaneously the distance residuals of points and plane matches in a non-linear least squares problem. Given two 3D point matches: $\{P,P'\}$ we express their residual in the vector form as: 

\begin{equation}
\label{eq:point_r}
\widetilde{P} = P' - (RP+t)
\end{equation}
whereas for two plane matches: $\{N,d\}$ and $\{N',d'\}$ we make use of the plane-to-plane distance, introduced in (\ref{eq:plane_dist}), such that, the residual can be derived, in the vector form, as: 
\begin{equation}
\label{eq:plane_r}
\widetilde{C} = (N'R(N't+d') - dN)^\top
\end{equation}
These residuals are then weighted, stacked together and minimized by using Levenberg-Marquart algorithm. More formally, we minimize the following joint cost function:
\begin{equation}
\label{eq:objective_fx}
\begin{aligned}
E = \sum_{i=1}^{N}w(\widetilde{P_i})\widetilde{P_i}^2 + \alpha\sum_{j=1}^{M}w(\widetilde{C_j})\widetilde{C_j}^2
\end{aligned}
\end{equation}
by iteratively recomputing the weights $w()$ based on the residual uncertainties, since these depend on the pose parameters. The residual uncertainties are computed using the first order error propagation of  (\ref{eq:point_r}) and (\ref{eq:plane_r}), given the pose estimate and the uncertainties of the point $\Sigma_P$ and plane extraction $\Sigma_{\theta}$ . Let $\Sigma_r$ represent the obtained uncertainty of the residual $r$ where $\{\sigma^2_1, \sigma^2_2,\sigma^2_3\}$ is the diagonal of $\Sigma_r$ then $w(r) = \begin{bmatrix}\sigma^{-2}_1 & \sigma^{-2}_2 & \sigma^{-2}_3 \end{bmatrix}$. Unlike the Mahalanobis distance, this weighting function neglects the covariances of $\Sigma_r$. Although, this is sub-optimal, in practice it is more efficient than using the Mahalanobis distance since it does not require inverting the 3$\times$3 residual uncertainties and it allows maintaining the residuals as vectors in the least squares problem, which we have found to inprove convergence. \par
Since the plane uncertainties depend on the weighting choice of the WLS plane fitting and planes are generally fewer than points, we introduce the scaling factor $\alpha$ in (\ref{eq:objective_fx}) to increase the impact of the planes on the optimization. Furthermore, an M-estimator with Tukey weights is used to further reweight the point residuals in order to down-weight the impact of outliers, whereas plane matching outliers are already addressed by the plane matching method (see Section \ref{matching}) and plane matches are too few to rely on statistics. \par
To avoid degenerate feature configurations, if the total number of point and plane matches is less than 3, pose optimization is avoided and instead a decaying velocity model \cite{PTAM} is applied. Additionally, after the optimization, the uncertainty of the pose parameters $\Sigma_{\xi}$ can be calculated since the Hessian evaluated at the solution is: $H=\Sigma_{\xi}^{-1}$. In this work, we use the Gauss-Newton approximation to the Hessian: $H \approx J^{\top}_rJ_r$, where $J_r$ is the combined Jacobian matrix of the residuals with respect to the pose parameters, and then validate the estimated parameters by checking the largest eigenvalue of the obtained $\Sigma_{\xi}$. If it is larger than a given threshold, we ignore the optimized pose and use the decaying velocity model instead.
\section{Experiments and Results}

\begin{figure}[t]
\centering
	\begin{tabular}{@{}c@{ }c@{}}
	\includegraphics[scale=0.5]{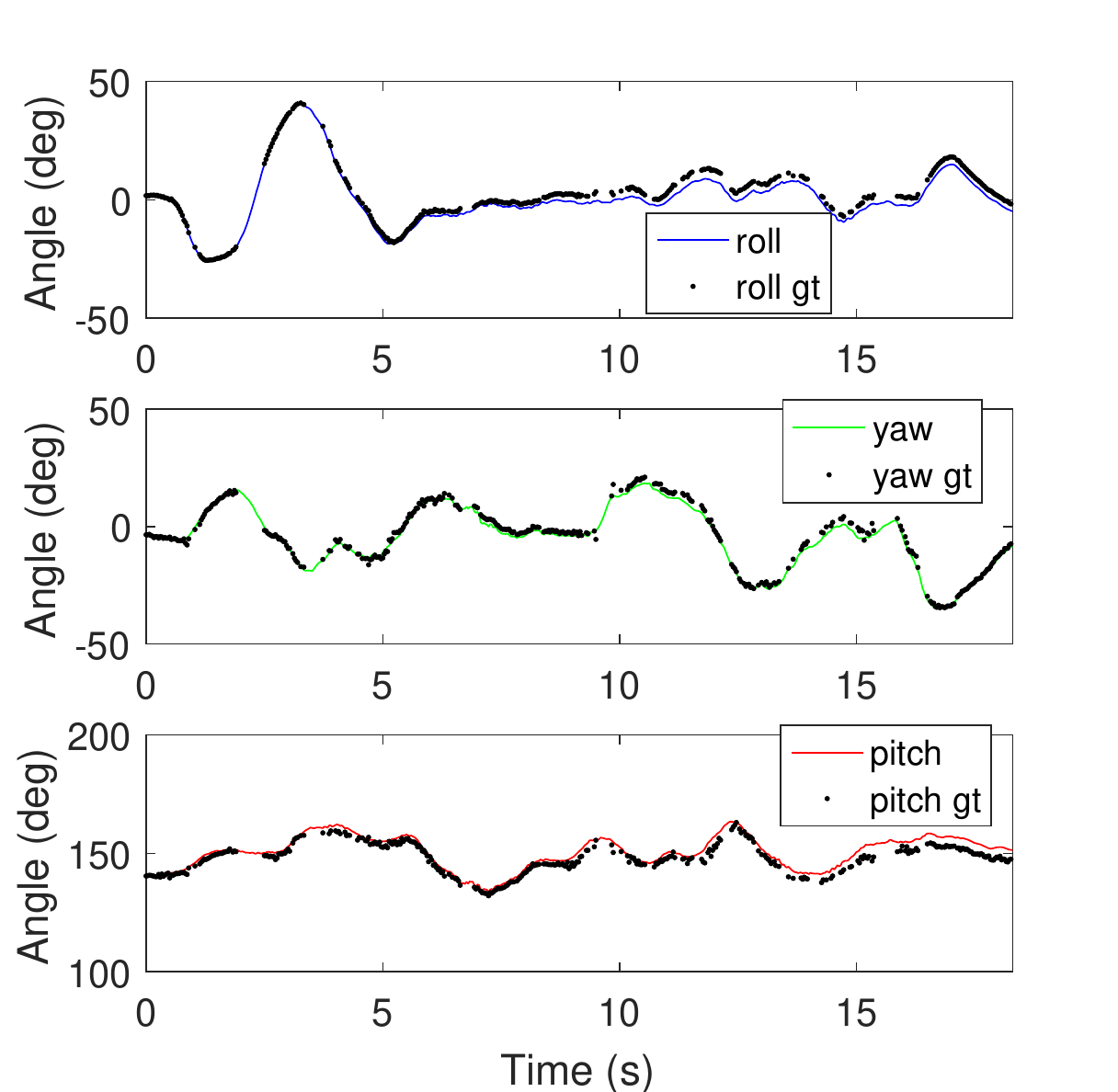}&
	\includegraphics[scale=0.5]{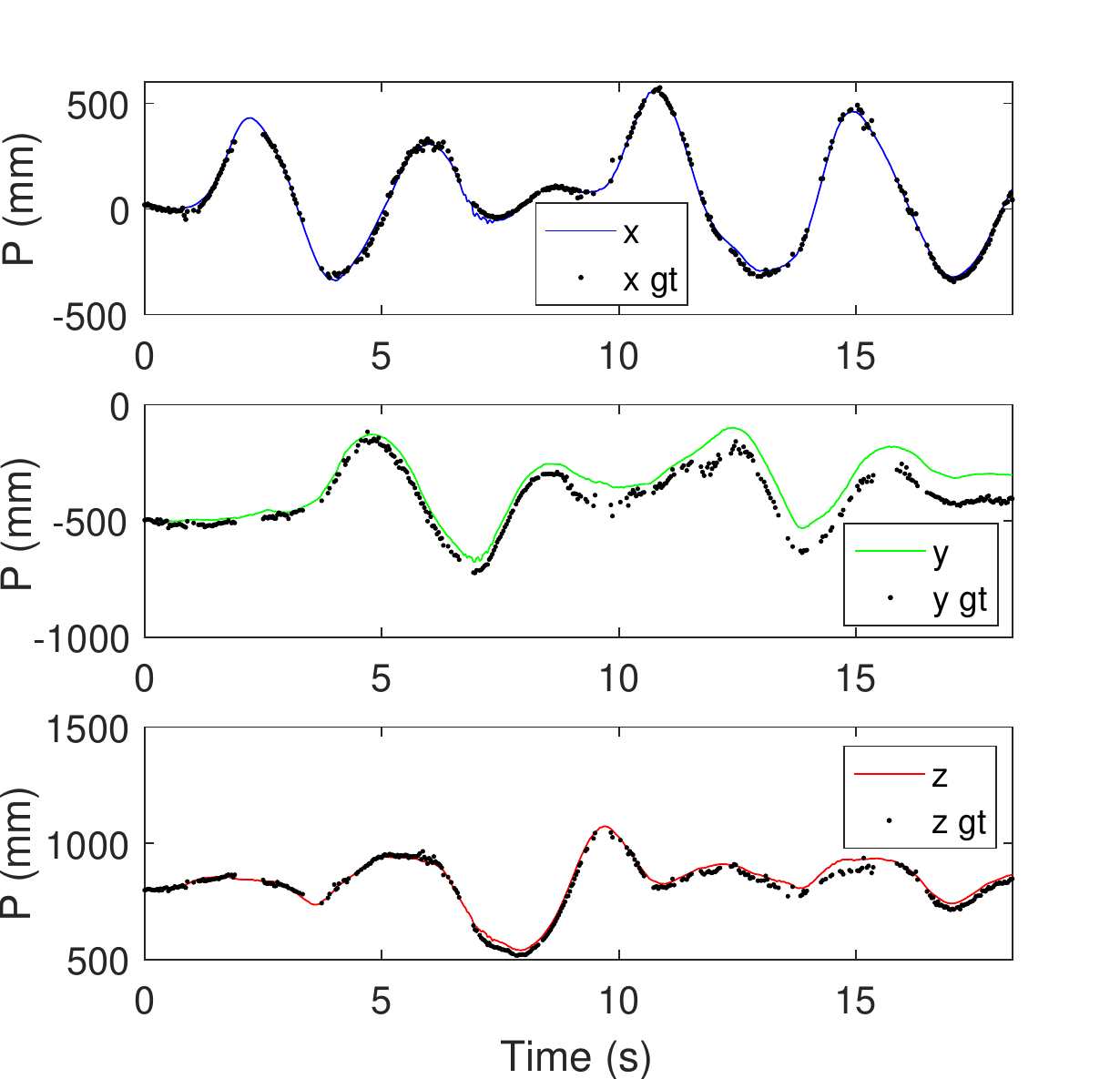}\\
	\multicolumn{2}{c}{\includegraphics[scale=0.30]{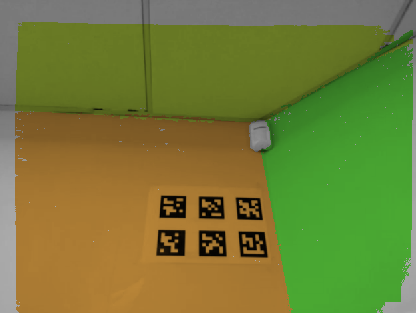}}
	\end{tabular}
	\caption{Top: Estimated pose for plane-based odometry on the corner sequence with ground-truth (gt) measured by using the markerboard. Bottom: Example from the corner sequence with detected planes overlayed on the color image.}
	\label{fig5}
\end{figure}

To validate the proposed visual odometry using plane primitives, we collected an RGB-D sequence with a Kinect sensor pointing towards a room corner, as shown in Fig. \ref{fig5}, for a corner is formed by three planes in a non-degenerate configuration. The camera was moved randomly in all 6 DoF, as shown in Fig. \ref{fig5}, during approximately 18 s. The final error of the position estimated by the plane odometry was around $114$ mm.

Moreover, the proposed method was evaluated on the TUM RGB-D dataset \cite{rgbd_dataset12iros}, which contains several RGB-D sequences along with the pose ground-truth, provided by a motion capture system. The sequences that were evaluated, in this work, are shown in Fig. \ref{fig4}. To assess the performance of the visual odometry, pose drift was measured as the relative pose error per second, as suggested by \cite{rgbd_dataset12iros}. \par

{\renewcommand{\arraystretch}{1.3}
\begin{table}[]
\centering
\caption{RMSE of relative pose per second for point-based odometry on TUM dataset. The probabilistic version corresponds to the proposed point odometry with uncertainty weighting, whereas the deterministic version uses unweighted residuals.}
\label{pt_odometry}
\begin{tabular}{|l|c|c|}
\hline
\multirow{2}{*}{Sequence} & \multicolumn{2}{c|}{Point Odometry} \\ \cline{2-3} 
 & Deterministic & Probabilistic \\ \hline
\multicolumn{1}{|c|}{\ fr1/desk} & \ \ \ 43 mm / 2.3\degree & \ 38 mm / 2.2\degree \\ \hline
\multicolumn{1}{|c|}{fr1/360} & \ 109 mm / 3.7\degree & \ 86 mm / 3.5\degree \\ \hline
\end{tabular}
\end{table}
}

\begin{figure}
\centering
	\includegraphics[scale=0.5]{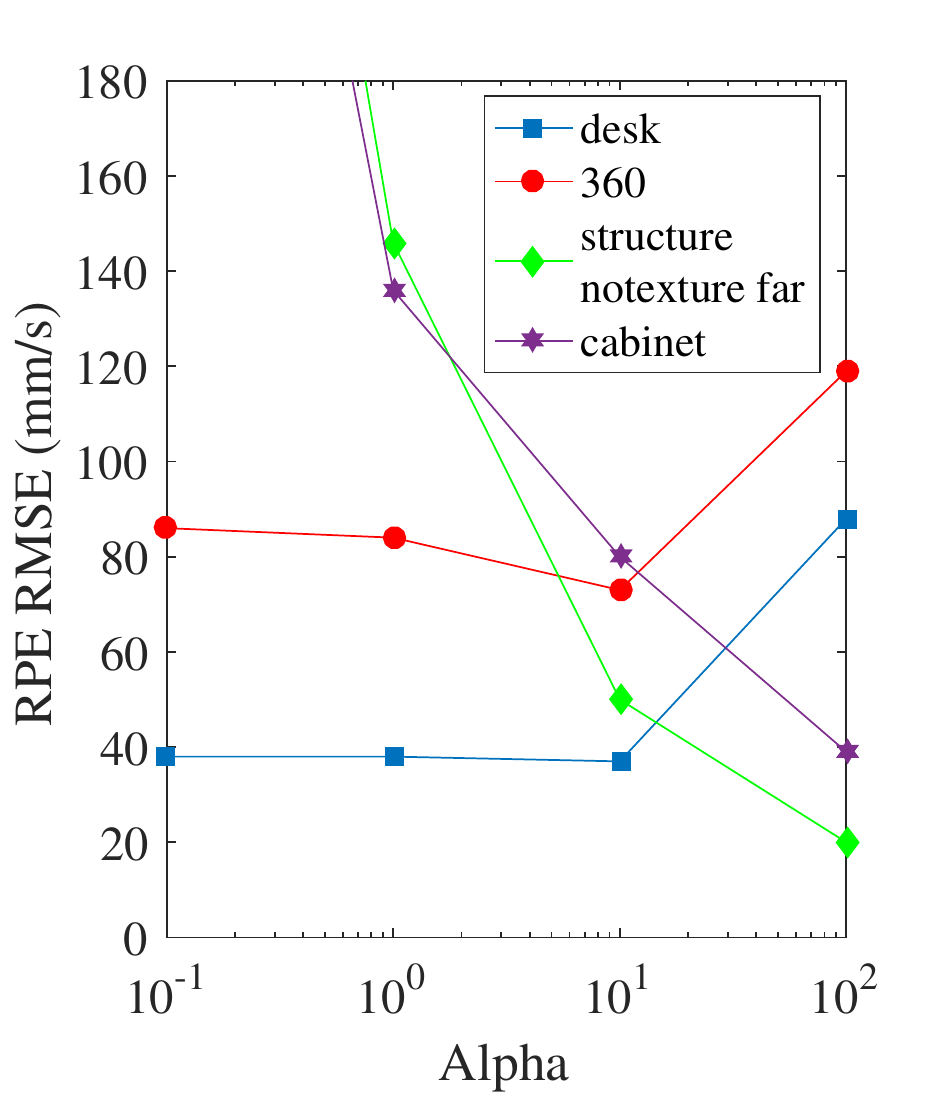}
	\caption{Alpha tuning. The higher the alpha, the more impact planes have on the pose estimation}
	\label{fig3}
\end{figure}

First, as can be observed in Table \ref{pt_odometry}, the visual odometry based on 3D-to-3D point matches is significantly improved by incorporating uncertainty in the pose estimation instead of relying simply on the Euclidean distance.
Fig. \ref{fig3} shows the impact of changing $\alpha$, i.e., the scale factor of the plane residuals, in the point and plane odometry, for each capture. Although, the performance on the `desk' sequence does not seem to benefit from the introduction of planes, in the low textured sequences, using planes proves to be advantageous. Furthermore, even though planes are not suited to be used alone, due to degeneracy, the curves of the low textured sequences indicate low error for high alpha. The contrary is observed for the other sequences, which suggests that better overall performance could be achieved by adjusting $\alpha$ dynamically based on the number of point matches. 
{\renewcommand{\arraystretch}{1.3}
\begin{table}[]
\centering
\caption{RMSE of relative pose per second for visual odometry with only points and with combination of points and planes. We also report, to the best of our knowledge, the best published relative translational error obtained by other frame-to-frame odometry methods that use neither loop closure detection nor map optimization. Since the method \cite{yang2017directLines} uses only a monocular camera, scale post-estimation was performed to best fit the groundtruth trajectory.}
\label{pt_ln_odometry}
\begin{tabular}{|l|c|c|c|}
\hline
\multirow{2}{*}{Sequence} & \multicolumn{2}{c|}{Features} & \multirow{2}{*}{State-of-the-art} \\ \cline{2-3} 
 & Points & Points \& Planes & \\ \hline
fr1/desk & \ 38 mm / 2.2\degree & \  37 mm / 2.1\degree & 26 mm \cite{gutierrez2016dense}\\ \hline
fr1/360 & \ 86 mm / 3.5\degree & \  73 mm / 2.9\degree & 84 mm \cite{PLVO}\\ \hline
fr3/structure\_notexture\_far \ & \ Fail & \  50 mm / 1.5\degree & 43 mm \cite{yang2017directLines}\\ \hline
fr3/cabinet & \ Fail & \ 80 mm / 4.0\degree & 133 mm \cite{yang2017directLines}\\ \hline
\end{tabular}
\end{table}
}

Nevertheless, the results for point and plane odometry, with the best overall tradeoff: $\alpha = 10$, are reported in Table \ref{pt_ln_odometry}. Although the sequences `desk' and `360' were captured in the same space, the sequence `360' was captured under more sudden rotations, which blurred the images, consequently yielding fewer good feature points, therefore using planes improved significantly the performance on that sequence.

\begin{figure}
\centering
	\begin{tabular}{@{}c@{ }c@{}c@{}}
		\includegraphics[scale=0.2]{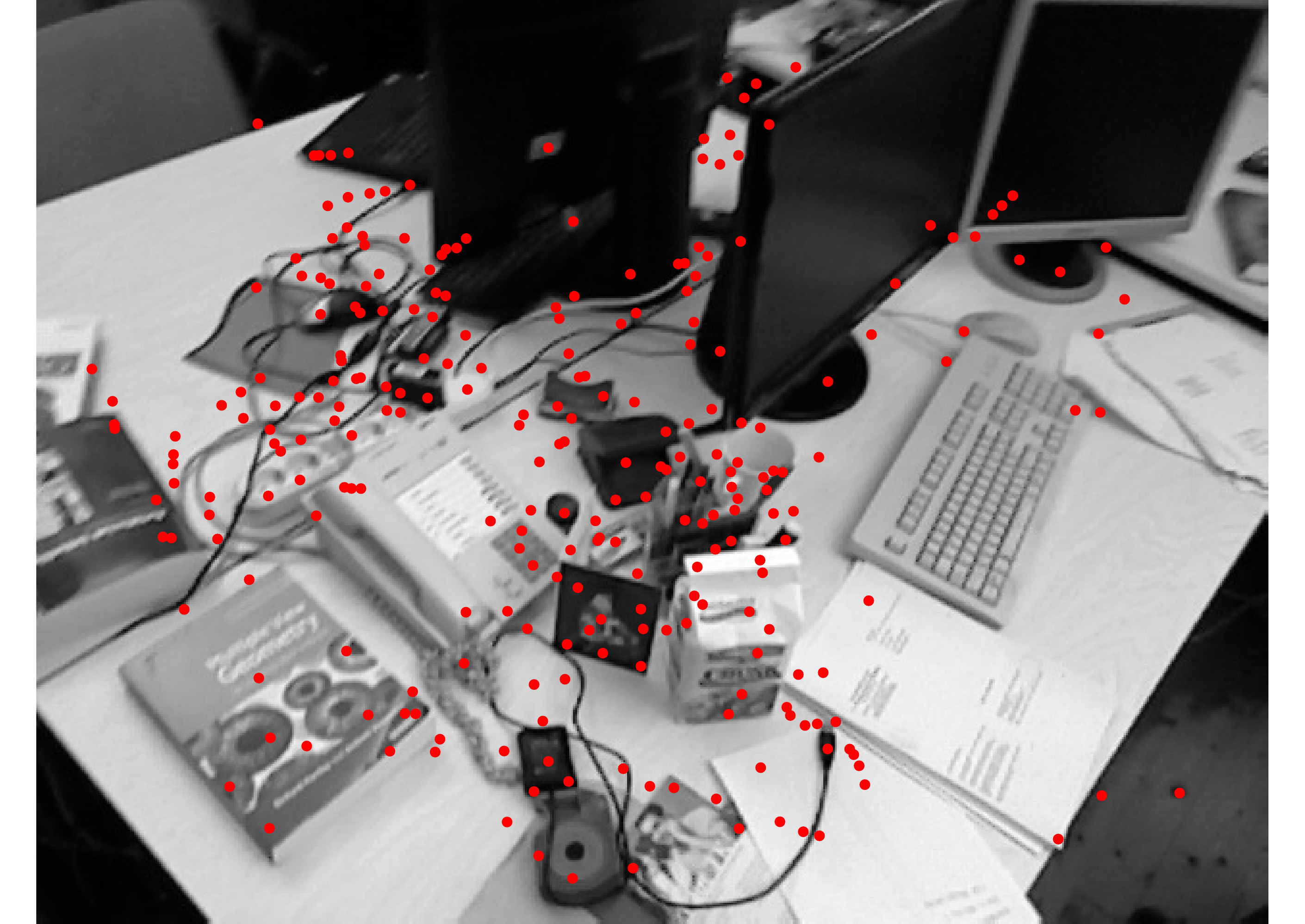} &
		\includegraphics[scale=0.2]{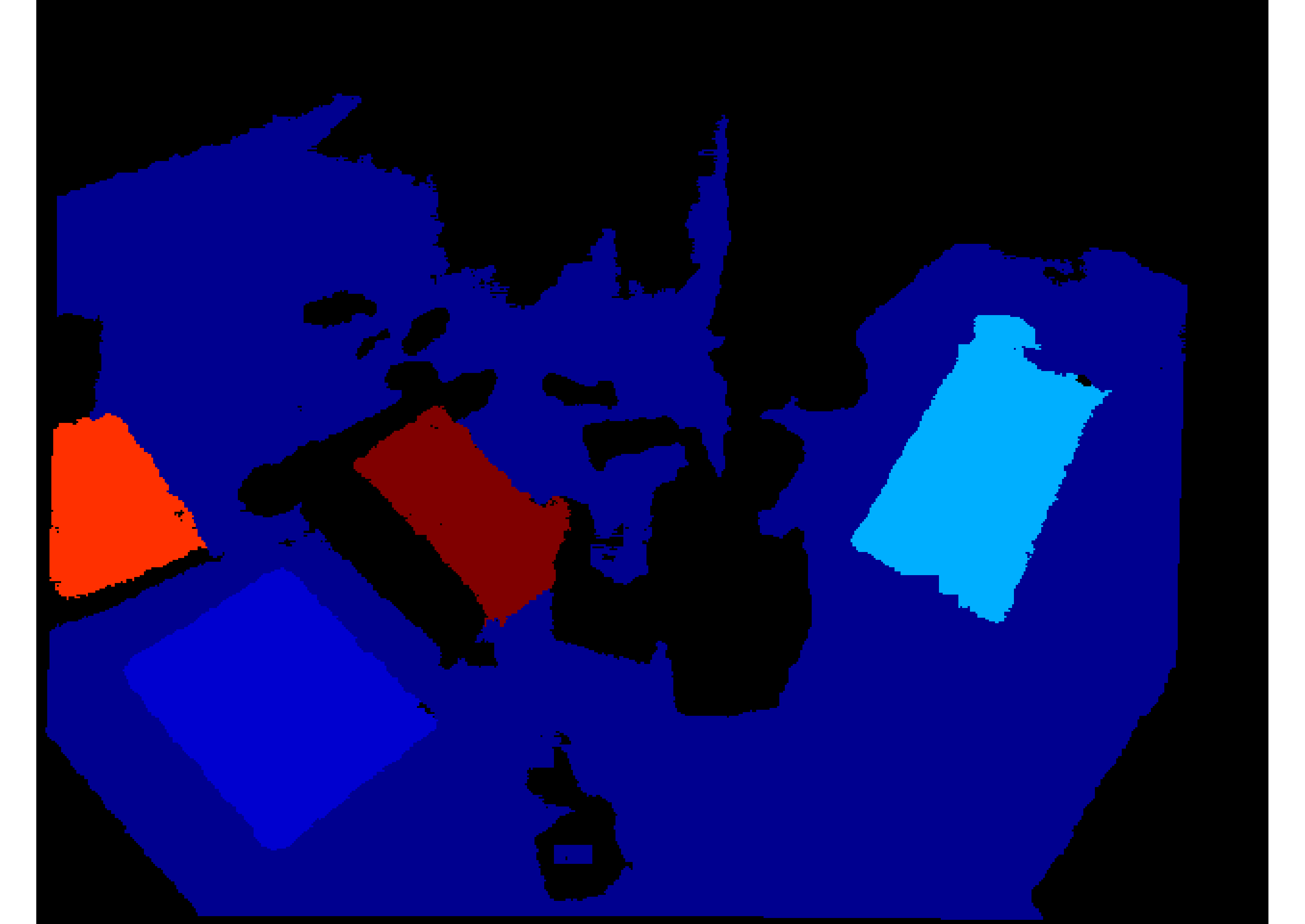}&
		\includegraphics[scale=0.13]{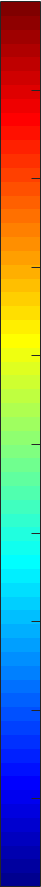}\\
		 \multicolumn{2}{c}{(a) fr1/desk} \\ \\
		\includegraphics[scale=0.2]{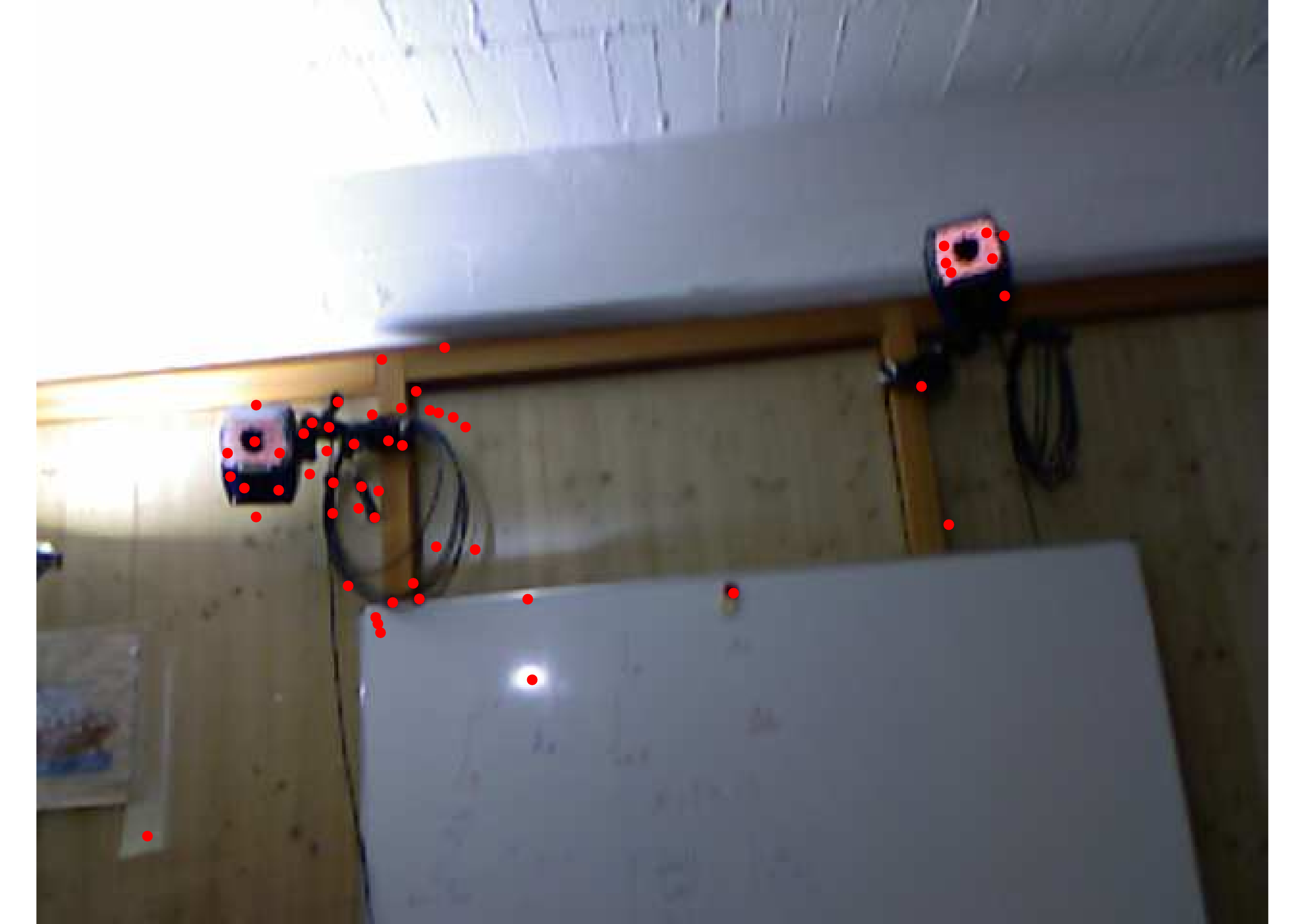} &
		\includegraphics[scale=0.2]{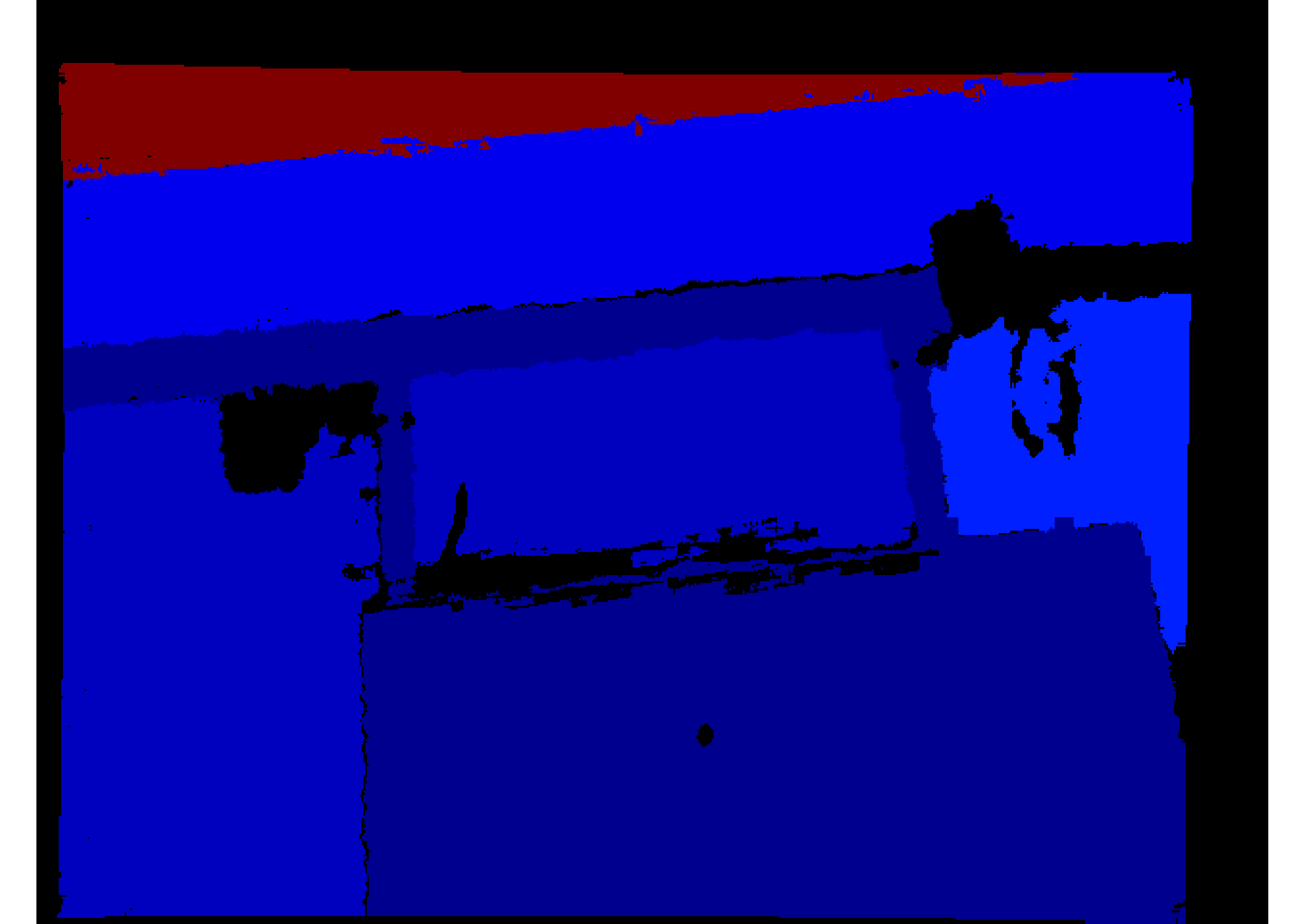}&
		\includegraphics[scale=0.13]{figs/color_bar.png}\\
		 \multicolumn{2}{c}{(b) fr1/360} \\ \\
		\includegraphics[scale=0.2]{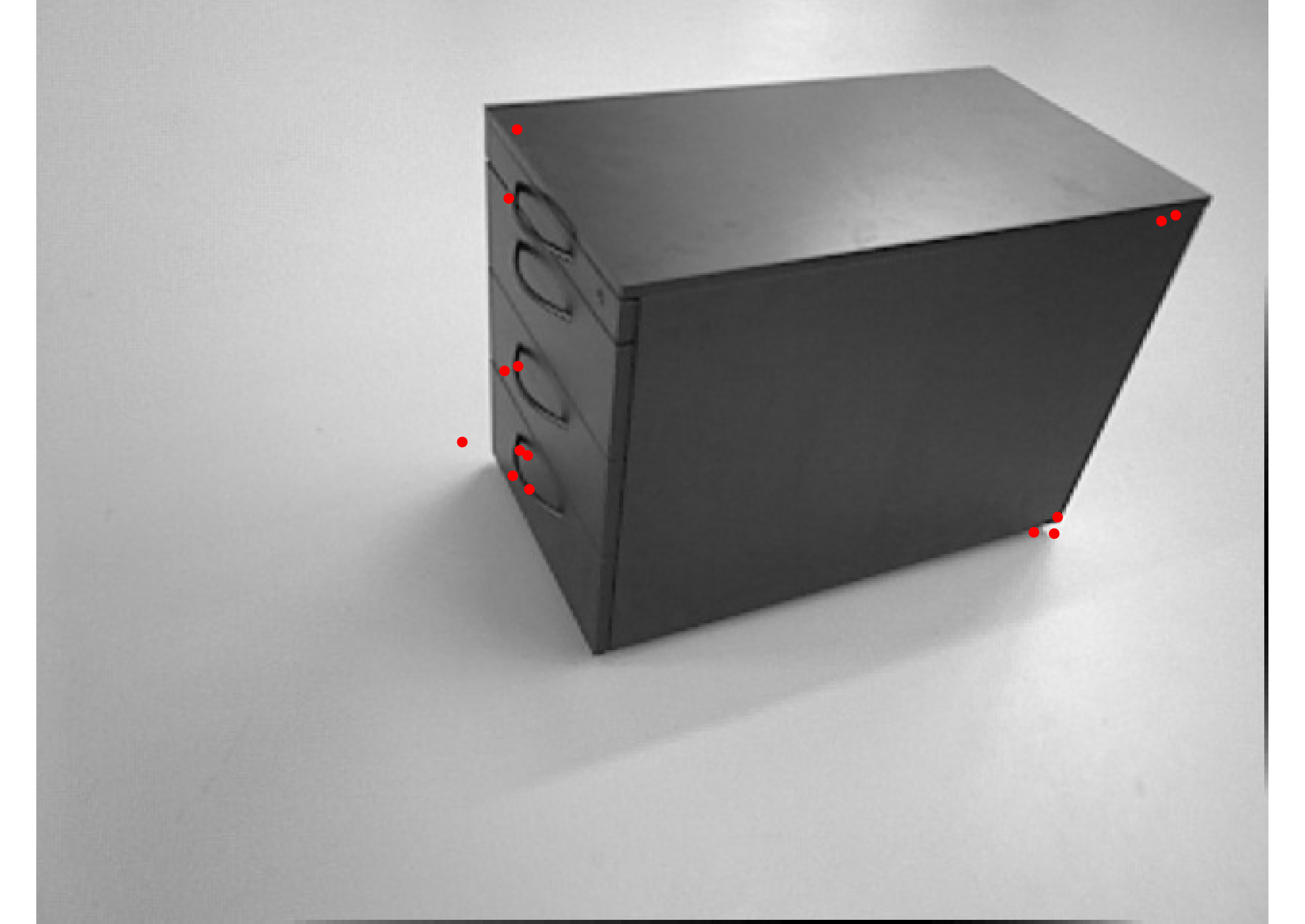} &
		\includegraphics[scale=0.2]{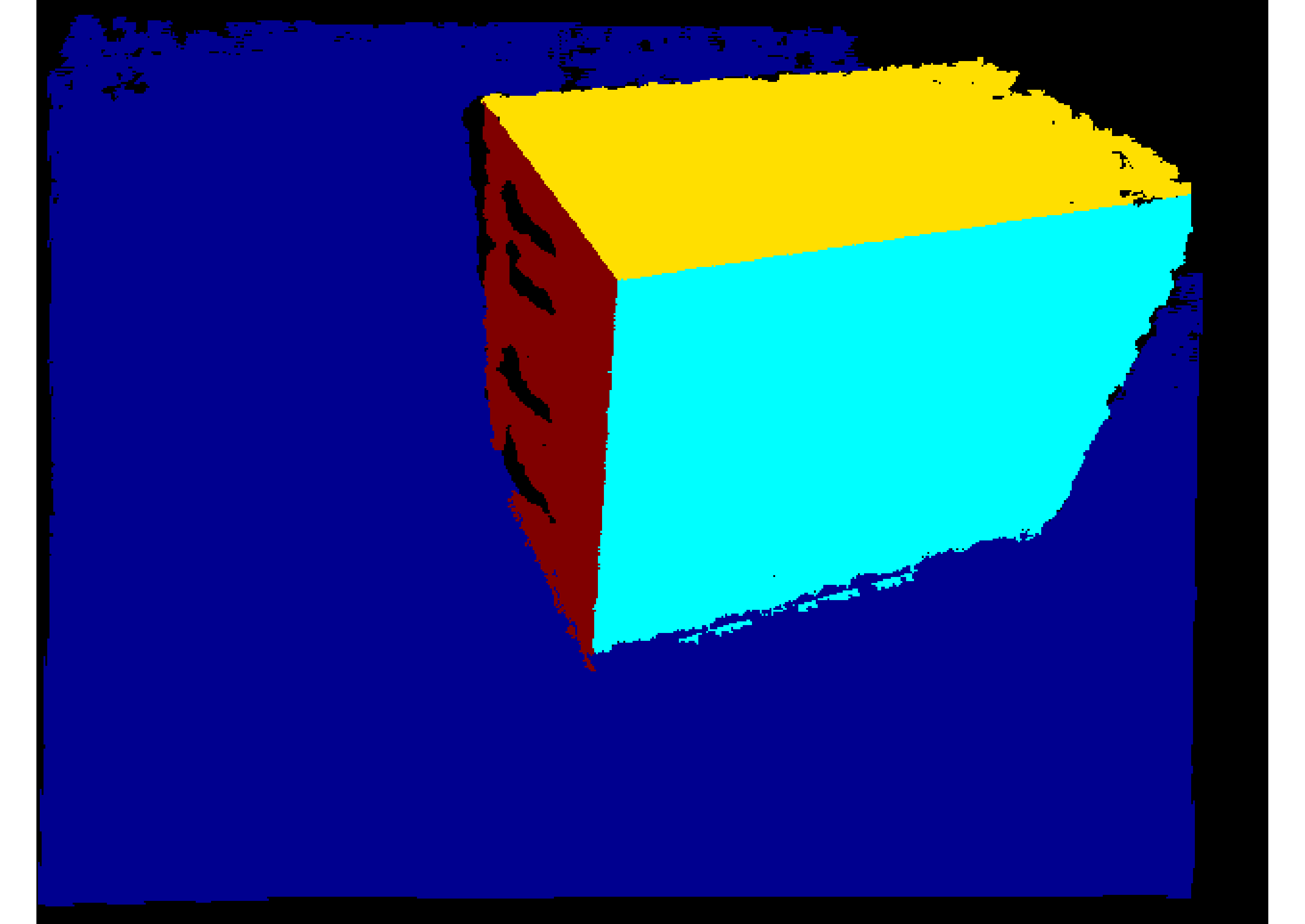} &
		\includegraphics[scale=0.13]{figs/color_bar.png}\\
		 \multicolumn{2}{c}{(c) fr3/cabinet} \\ \\
		 \includegraphics[scale=0.2]{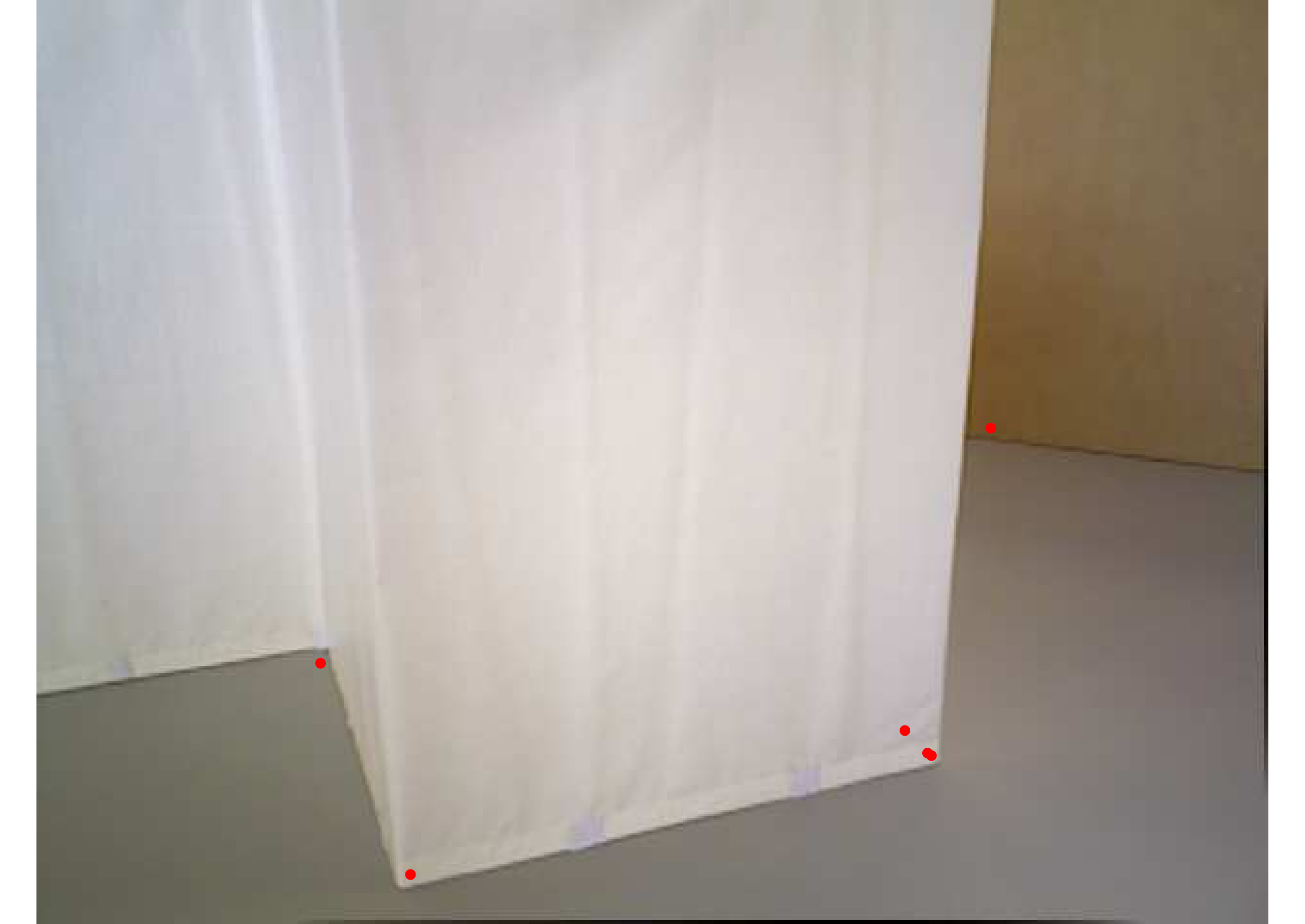} &
		 \includegraphics[scale=0.2]{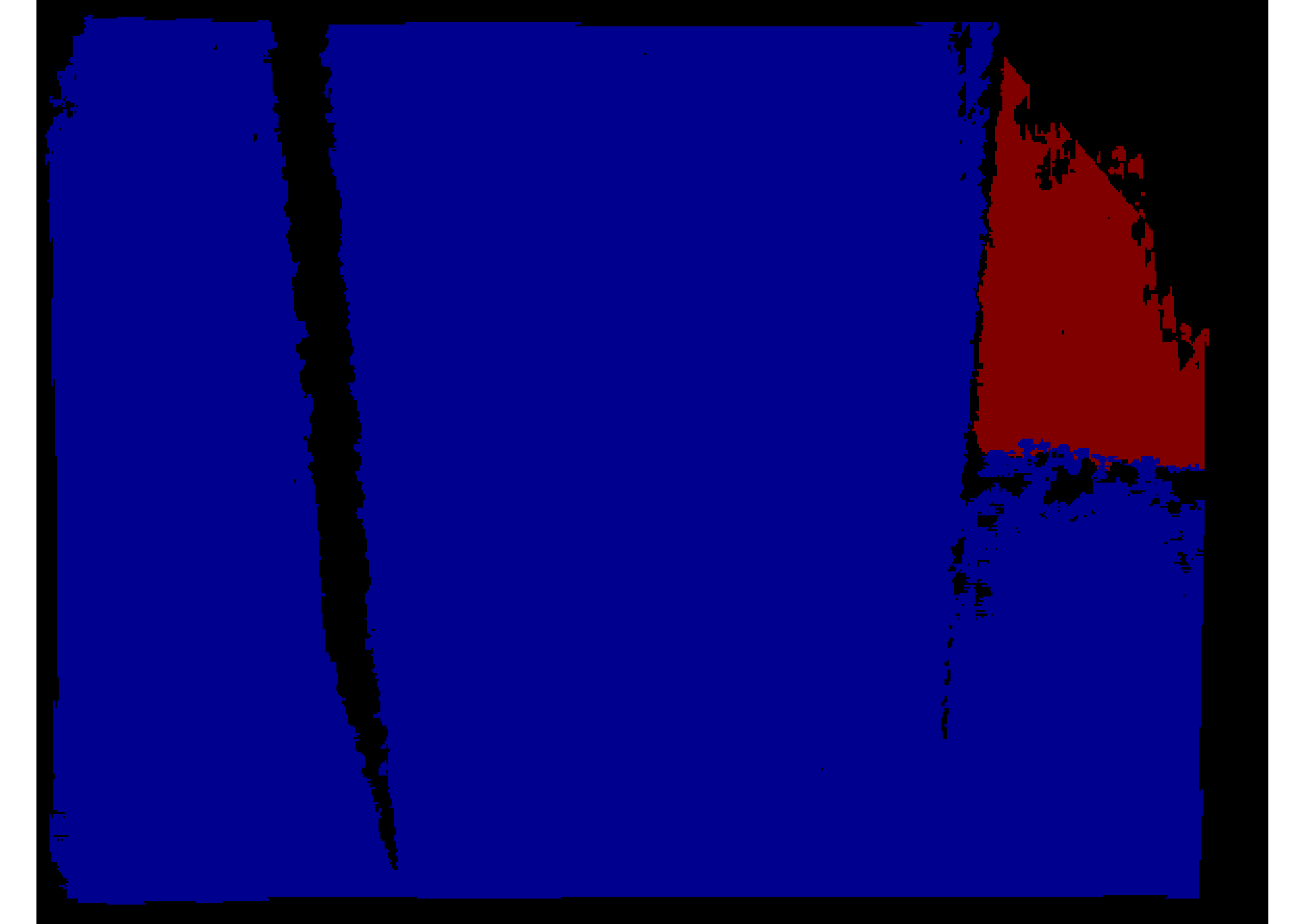} &
		 \includegraphics[scale=0.13]{figs/color_bar.png}\\
		 \multicolumn{2}{c}{(c) fr3/structure\_notexture\_far}
	\end{tabular} 
	\caption{Examples of frames from the evaluated dataset sequences. Left: Detected image points. Right: Detected planes colored by the normalized uncertainty of their estimated distance to the origin}
	\label{fig4}
\end{figure}

\section{Conclusion and Future Work}
This paper presents a visual odometry method that combines points and planes based on their measurement uncertainties. Our results demonstrate that this feature combination is beneficial, when few image feature points are detected either due to non-textured planar surfaces or blur caused by sudden motion. \par Furthermore, results of point-based odometry show that the systematic noise of a structured-light camera affects significantly the pose estimation based on 3D-to-3D alignment, since our weighing method leads to superior results. Therefore working on the Euclidean space \cite{PointsAndPlanes_2013} should be avoided when using such sensors. Alternatively, using the reprojection error has also been advocated \cite{RGBD2012henry}, as it is less susceptible to depth errors, however the intent of this work was to demonstrate the effect of modelling depth uncertainty in the worst case. \par
As future work, we plan to extend the framework to line features and investigate more comprehensive uncertainty models, such as \cite{dryanovski2013fast}, which addresses depth discontinuities, unlike the model used in this work.

\subsubsection*{Acknowledgments.} This work was supported by Sellafield Ltd.

\bibliographystyle{splncs03}
\bibliography{taros_ref}

\begin{thebibliography}{10}
\providecommand{\url}[1]{\texttt{#1}}
\providecommand{\urlprefix}{URL }

\bibitem{bay2006surf}
Bay, H., Tuytelaars, T., Van~Gool, L.: Surf: Speeded up robust features. In:
  Computer vision--ECCV, pp. 404--417. Springer (2006)

\bibitem{dryanovski2013fast}
Dryanovski, I., Valenti, R.G., Xiao, J.: Fast visual odometry and mapping from
  rgb-d data. In: IEEE International Conference on Robotics and Automation
  (ICRA). pp. 2305--2310 (2013)

\bibitem{feng2014fast}
Feng, C., Taguchi, Y., Kamat, V.R.: Fast plane extraction in organized point
  clouds using agglomerative hierarchical clustering. In: IEEE International
  Conference on Robotics and Automation (ICRA). pp. 6218--6225 (2014)

\bibitem{gutierrez2016dense}
Gutierrez-Gomez, D., Mayol-Cuevas, W., Guerrero, J.J.: Dense rgb-d visual
  odometry using inverse depth. Robotics and Autonomous Systems  75,  571--583
  (2016)

\bibitem{RGBD2012henry}
Henry, P., Krainin, M., Herbst, E., Ren, X., Fox, D.: Rgb-d mapping: Using
  kinect-style depth cameras for dense 3d modeling of indoor environments. The
  International Journal of Robotics Research  31(5),  647--663 (2012)

\bibitem{kaess2015simultaneous}
Kaess, M.: Simultaneous localization and mapping with infinite planes. In: IEEE
  International Conference on Robotics and Automation (ICRA). pp. 4605--4611
  (2015)

\bibitem{Khoshelham2012}
Khoshelham, K., Elberink, S.O.: {Accuracy and resolution of kinect depth data
  for indoor mapping applications}. Sensors  12(2),  1437--1454 (2012)

\bibitem{PTAM}
Klein, G., Murray, D.: Parallel tracking and mapping for small ar workspaces.
  In: IEEE and ACM International Symposium on Mixed and Augmented Reality
  (ISMAR). pp. 225--234 (2007)

\bibitem{PLVO}
Lu, Y., Song, D.: Robust rgb-d odometry using point and line features. In: IEEE
  International Conference on Computer Vision (ICCV) (2015)

\bibitem{ma2016cpa}
Ma, L., Kerl, C., St{\"u}ckler, J., Cremers, D.: Cpa-slam: Consistent
  plane-model alignment for direct rgb-d slam. In: IEEE International
  Conference on Robotics and Automation (ICRA). pp. 1285--1291 (2016)

\bibitem{pathak2010uncertainty}
Pathak, K., Vaskevicius, N., Birk, A.: Uncertainty analysis for optimum plane
  extraction from noisy 3d range-sensor point-clouds. Intelligent Service
  Robotics  3(1),  37--48 (2010)

\bibitem{PlaneAndsurfels_2014}
Salas-Moreno, R.F., Glocken, B., Kelly, P.H., Davison, A.J.: Dense planar slam.
  In: IEEE International Symposium on Mixed and Augmented Reality (ISMAR). pp.
  157--164 (2014)

\bibitem{rgbd_dataset12iros}
Sturm, J., Engelhard, N., Endres, F., Burgard, W., Cremers, D.: A benchmark for
  the evaluation of rgb-d slam systems. In: International Conference on
  Intelligent Robots and Systems (IROS) (2012)

\bibitem{PointsAndPlanes_2013}
Taguchi, Y., Jian, Y.D., Ramalingam, S., Feng, C.: Point-plane slam for
  hand-held 3d sensors. In: IEEE International Conference on Robotics and
  Automation (ICRA). pp. 5182--5189 (2013)

\bibitem{trevor2012planar}
Trevor, A.J., Rogers, J.G., Christensen, H.I.: Planar surface slam with 3d and
  2d sensors. In: IEEE International Conference on Robotics and Automation
  (ICRA). pp. 3041--3048 (2012)

\bibitem{weingarten2004probabilistic}
Weingarten, J.W., Gruener, G., Siegwart, R.: Probabilistic plane fitting in 3d
  and an application to robotic mapping. In: IEEE International Conference on
  Robotics and Automation (ICRA). vol.~1, pp. 927--932 (2004)

\bibitem{yang2017directLines}
Yang, S., Scherer, S.: Direct monocular odometry using points and lines. arXiv
  preprint arXiv:1703.06380  (2017)

\end{thebibliography}

\end{document}